\documentclass{infor}
\volume{0}
\issue{0}
\pubyear{2022}
\articletype{research-article}
\doi{0000}

\theoremstyle{remark}

\begin{document}
\begin{frontmatter}
\pretitle{Research Article}

\title{WISCA: A Consensus-Based Approach to Harmonizing Interpretability in Tabular Datasets}
\runtitle{WISCA: Interpretable Consensus for Tabular Data}

\author{\inits{A.}\fnms{Antonio Jesús} \snm{Banegas-Luna}\thanksref{c1}\ead[label=e1]{ajbanegas@ucam.edu}\bio{bio1}}
\author{\inits{H.}\fnms{Horacio} \snm{Pérez-Sánchez}\ead[label=e1]{hperez@ucam.edu}\bio{bio2}}
\author{\inits{C.}\fnms{Carlos} \snm{Martínez-Cortés}\ead[label=e1]{cmartinez1@ucam.edu}\bio{bio3}}
\thankstext[type=corresp,id=c1]{Corresponding author.}
\address{Campus Los Jerónimos s/n 30107 Guadalupe (Murcia), \institution{Universidad Católica de Murcia (UCAM)}, \cny{Spain}}


\begin{abstract}
While predictive accuracy is often prioritized in machine learning (ML) models, interpretability remains essential in scientific and high-stakes domains. However, diverse interpretability algorithms frequently yield conflicting explanations, highlighting the need for consensus to harmonize results. In this study, six ML models were trained on six synthetic datasets with known ground truths, utilizing various model-agnostic interpretability techniques. Consensus explanations were generated using established methods and a novel approach: WISCA (Weighted Scaled Consensus Attributions), which integrates class probability and normalized attributions. WISCA consistently aligned with the most reliable individual method, underscoring the value of robust consensus strategies in improving explanation reliability.
\end{abstract}


\end{frontmatter}

\section{Introduction}

Machine learning (ML), a key area of artificial intelligence, enables computers to learn autonomously from data. The robust statistical foundation of ML has facilitated its adoption for intricate tasks in various scientific domains, from biology \cite{Qu2019, Jones2019} to drug discovery \cite{Vamathevan2019, Ekins2019, Elbadawi2021}, meteorology \cite{Bi2023, Pathak1001, Stirnberg2021}, and, notably, medicine \cite{Waljee2010, Sidey2019, Rajkomar2019}, due to its precision in unveiling patterns within complex datasets.

Although ML models are typically evaluated based on predictive accuracy, transparency in their decision-making is increasingly demanded, especially in sensitive domains such as healthcare and finance. The opaqueness of certain models is particularly problematic in sectors where understanding model logic is as critical as the outcomes themselves, such as healthcare \cite{Yang2022,Payrovnaziri2020,Zhang2022} and financial services \cite{Chen2023,Demajo2020,Vcernevivciene2022}. This imperative for clarity has catalyzed the emergence of eXplainable Artificial Intelligence (XAI) \cite{Gilpin2018,bennetot_practical_2024}, an initiative to unveil the rationale behind model predictions. In addition, the ability to explain predictions is becoming a legal and ethical requirement in some jurisdictions, particularly in high-stakes domains such as healthcare and credit scoring \cite{Pavlidis02012024}.

Interpretability algorithms, designed to elucidate ML models, are classified as global or local based on their explanatory scope or by their specificity to model types versus model-agnostic versatility \cite{Linardatos2020,Molnar2020}. As model accuracy is typically assessed using metrics such as AUC, Precision, Recall, R$^2$, MSE, and MAE \cite{Rainio2024,Zhou2021}, the quality of explanations is evaluated through criteria like correctness, consistency, coherence, and confidence \cite{Nauta2023}. These factors play a crucial role in determining the effectiveness and reliability of the explanations provided by AI systems. Researchers have developed different scales and metrics to measure the quality of explanations, such as the Explanation Satisfaction Scale, Explanation Goodness Checklist, and the System Causability Scale \cite{Hoffman2023}. The Explanation Satisfaction Scale focuses on users' evaluations of explanations, while researchers use the Explanation Goodness Checklist to independently assess the quality of explanations. On the other hand, the System Causability Scale aims to measure the quality of explanations in explainable AI systems, considering aspects like timeliness, detail, completeness, understandability, and learnability of the explanation \cite{Holzinger2020}.

However, despite being metrics widely adopted by the community, they usually refer to subjective concepts that are difficult to quantify. In this regard, progress is being made in the search for objective metrics to assess the quality of the explanations generated by interpretability algorithms \cite{Islam2020,Rosenfeld2021}. However, the utility of XAI is often undercut by the inconsistent explanations these algorithms generate, leading to confusion and diminishing trust \cite{Rudin2022, Zhou2021, Krishnan2020, Vowels2022}. This disagreement is particularly pronounced when different interpretability methods are applied to the same prediction and produce contradictory results, leaving the end users without a clear understanding of the behavior of the model \cite{Krishnan2020}. The challenge, therefore, lies in synthesizing these divergent narratives into a coherent consensus that accurately reflects the collective wisdom of multiple algorithms. Yet, there is no established framework for systematically comparing or integrating these explanations in a reliable way \cite{SAEED2023110273}.

To this end, consensus functions have emerged as a solution to harmonize explanations, with methods ranging from the arithmetic mean to other statistical measures and feature occurrence-based approaches \cite{Sarlette2009, Zhang2019, Bajusz2019, Burgos2019, Ayad2007, Ayad2010, Fischman2011}. Yet, these functions often fall short by not accounting for factors such as model accuracy, which can significantly influence the reliability of the derived explanations. Incorporating model accuracy into the consensus functions can enhance the robustness and trustworthiness of the explanations provided by AI systems. By considering model accuracy, the resulting explanations can better reflect the true behavior and decision-making processes of the underlying AI models, leading to more reliable and informative insights for users. This holistic approach to explanation harmonization can contribute to improving the overall transparency and interpretability of AI systems, fostering greater trust and understanding among users and facilitating more effective decision-making processes. Moreover, better consensus mechanisms could directly benefit clinical and industrial stakeholders by reducing uncertainty and enabling auditable explanation workflows \cite{lekadir2024futureaiinternationalconsensusguideline}.

This study explores several consensus functions applied to ML models, identifies limitations of traditional approaches, and introduces a new method, WISCA, that incorporates class probabilities and attribution scaling for a more robust explanation process. Our goal is to bridge the gap between model performance and explanation quality by prioritizing both in the consensus process. The subsequent sections detail the models, datasets, interpretability algorithms, and consensus functions utilized in our study, with Section \ref{sec:materials} presenting the methodology, Section \ref{sec:results} presenting the findings, and Section \ref{sec:discussion} delving into a comprehensive evaluation of the consensus functions. Ultimately, we consolidate our key insights and propose future research trajectories.

\section{Materials and methods}
\label{sec:materials}
This section describes the synthetic datasets created for the experiments, the set of ML models used, the interpretability algorithms used to explain the internal workings of the models, and the consensus functions assessed. Finally, a novel consensus function is proposed to mitigate the identified challenges. 

\subsection{Datasets}

In order to analyze disagreements in classification and regression problems, six synthetic datasets were created: two for binary classification problems, two for multiclass classification problems, and two for regression problems. The main reason for choosing synthetic over toy datasets is that they can be constructed using predefined rules based on specific input features. This facilitates the evaluation of whether the resulting explanations correctly identify the expected features, since the features expected to explain the model are known in advance.

In all datasets, the target feature was calculated based on 3 or 4 input features by applying different linear and non-linear functions. In addition, several irrelevant features that simulate noise were added to the datasets. This was done to verify whether the interpretability algorithms focused on the most relevant features or whether they were disturbed by the noise instead.

The datasets contained between 20 and 75 features. The number of features in each data set was chosen randomly so that the variables not involved in the target computation were sufficient to prevent the model from learning the target feature too easily. Therefore, each dataset includes 16 to 72 noise variables designed to reduce model overfitting. Furthermore, to avoid time-consuming calculations, since each model was trained 10 times, each dataset was created with a random number of samples ranging from 1500 to 2500. With this number of samples, the models have enough data to learn effectively and can be trained in a few minutes or hours. All features were generated using a uniform distribution between 0 and 1. In this way, collections of normalized datasets were quickly simulated.

Concerning the formulas used to model the target variables, linear and non-linear functions have been used to generate different types of dataset. 


Table \ref{tab:datasets} summarizes the main properties of each dataset and the combination of variables implemented to calculate the target. For each dataset, its name, the type of task it solves, the number of samples, and features are indicated. The \textit{Expected Explanation} column lists the features from which the target variable was calculated. Therefore, these are the variables that the interpretability algorithms were expected to identify. Finally, Table \ref{tab:formulas} shows the formula implemented to calculate the output values.

\begin{table}
\caption{Description of the synthetic datasets used in this work.}\label{tab:datasets}
\begin{tabular*}{\textwidth}{@{\extracolsep{\fill}}cccccrrrr@{}}
\hline
Dataset & Type & \# Samples & \# Features & Expected Explanation$^1$\\
\hline
Dataset 1 &Binary &2000 &20  &F2, F3, F9, F17 \\
Dataset 2 &Binary &1500 &75 &F5, F25, F55 \\
Dataset 3 &Regression &2500  &60  &F1, F56, F58, F60 \\
Dataset 4 &Regression &2000 &30 &F19, F21, F24, F26 \\
Dataset 5 &Multiclass &1000 &10 &F3, F4, F7, F10 \\
Dataset 6 &Multiclass &2500  &30  &F12, F16, F22, F27 \\
\hline
\end{tabular*}
\end{table}

\begin{table}[]
\caption{Formulas implemented in the datasets to calculate the target.}\label{tab:formulas}
\begin{tabular*}{\textwidth}{@{\extracolsep{\fill}}ll@{}}
\hline
Dataset   & Formula \\
\hline
Dataset 1 & if (F2*F3)/F9 \textless F17 then 0 else 1 \\
Dataset 2 & if (F55$^{3}$ + F5$^{2}$ - F25 \textless 0) then 0 else 1 \\
Dataset 3 & sin(F60) + cos(F58) + tanh(F56) + F1 \\
Dataset 4 & F19$^{4}$ - F21$^{3}$ + F24$^{2}$ - F26 \\
Dataset 5$^1$ & if\ (X \textless -4)\ then\ 0\ elsif\ (X $\in$ {[}-4, 0))\ then\ 1\ elsif\ (X $\geq$ 0)\ then\ 2 \\
Dataset 6$^2$ & \begin{tabular}[c]{@{}l@{}}if (X $\leq$ -12) then 0 elsif (X $\in$ (-12,-6{]}) then 1 elsif (X $\in$ (-6,0{]}) then 2 \\ elsif (X $\in$ (0,6{]}) then 3 elsif (X $\in$ (6,14{]}) then 4 elsif (X \textgreater 14) then 5\end{tabular} \\
\hline
\end{tabular*}
\footnote[1] \\X = (F7 * 2) + (F3 / 3) + F4 - F10 \\
\footnote[2] \\X = (F27 * 13) - F22 + F16$^2$ - (F12 * 1.5)
\end{table}

\subsection{Machine learning models}

The algorithms analyzed were used to interpret the predictions of different ML models. The selected models include k-nearest neighbors (KNN), random forests (RF), support vector machines (SVM), extreme gradient boosting machines (XGB), and artificial neural networks (ANN). They are a representative collection that includes non-linear, ensemble, and black-box models. In addition, a linear (logistic) regression model (LR) was added to be used as a baseline. With this approach, in this study a diverse set of model architectures is covered.

\subsection{Interpretability algorithms}

A representative set of interpretability algorithms was chosen for this work. Both global and local model-agnostic approaches were taken into consideration. Model-agnostic algorithms are applicable to any model architecture. Furthermore, it is worth noticing that global methods represent the contribution of a feature to model decisions with a single numerical value. This value is commonly referred to as \emph{attribution}. However, local approaches return an individual attribution of each feature by sample.

In this work, we used Random Forests \cite{Breiman2001} to estimate the importance of each input feature in the model's decision-making process. Specifically, we relied on the importance of impurity-based features, also known as mean decrease in impurity (MDI), as implemented in the Scikit-learn library. This method calculates the importance of a feature by measuring the total decrease in node impurity (e.g., Gini impurity) brought about by splits involving that feature, averaged over all trees in the forest. A higher decrease indicates a more informative feature. Furthermore, we used the importance of permutation-based features \cite{Altmann2010} as a complementary method. This approach evaluates the decrease in model performance when the values of a single feature are randomly permuted, effectively breaking its relationship to the target. Together, these methods provided a global view of the relative contribution of each feature to the predictive performance of the model.

On the other hand, Local Interpretable Model-agnostic Explanations (LIME) \cite{Ribeiro2016}, SHAP \cite{Lundberg2017, Strumbelj2014}, integrated gradients \cite{Sundararajan2017} and counterfactual explanations \cite{Mothilal2020} were selected as local approaches. 

\subsection{Consensus functions}

This work assessed different consensus functions previously published in the literature. Those functions will be the basis for developing and evaluating a novel consensus approach. The consensus functions evaluated in this study are described below.

\paragraph{Arithmetic mean} The arithmetic mean (Eq. \ref{eq:arithmetic_mean}) is a simple approach to average the importance of all the features. It is frequently used because of its simplicity of calculation. In addition, it is a fair approach that gives the same importance to all the explanations. However, the main problem with this function is that it weights equally the explanations of very precise and \textit{random} classified samples. Furthermore, since interpretability algorithms may produce attributions on different scales, applying the arithmetic mean to unnormalized attributions can bias the consensus.

\begin{equation}
\label{eq:arithmetic_mean}
A_{mean} = \frac{\sum_{i=1}^{n}{x_{i}}}{n}
\end{equation}

\paragraph{Harmonic mean} The harmonic mean is the reciprocal of the arithmetic mean (Eq. \ref{eq:harmonic_mean}). It is often used to average the data inversely proportional to the data. For example, calculate the average velocities to account for the effect of distance over time. Although it cannot deal with null or negative values, it is robust against extreme positive outliers. While the harmonic mean cannot handle zero or negative values, it can mitigate the influence of extreme outliers when the attribution scales vary.  

\begin{equation}
\label{eq:harmonic_mean}
H_{mean} = \frac{n}{\sum_{i=1}^{n}\frac{1}{x_{i}}}
\end{equation}

\paragraph{Geometric mean} The geometric mean can be expressed in terms of the arithmetic and the harmonic means (Eq. \ref{eq:geometric_mean_1}), but it can also be formulated in terms of individual attributions of each feature (Eq. \ref{eq:geometric_mean}). The geometric mean is often used to analyze time series and growth rates, where values are multiplicatively related to each other. It is less sensitive to extreme values than the arithmetic mean, but its calculation is more complex. Additionally, as well as the harmonic mean, it does not handle non-positive values correctly. Despite all those limitations, the geometric mean lies between the arithmetic and harmonic mean. Hence, it can be worth testing its performance as a consensus function.

\begin{equation}
\label{eq:geometric_mean_1}
G_{mean} = \sqrt{A_{mean}*H_{mean}}
\end{equation}

\begin{equation}
\label{eq:geometric_mean}
G_{mean} = \sqrt[n]{\prod_{i=1}^{n}x_{i}}
\end{equation}

\paragraph{Voting} A limitation due to the usage of attribution scores is that not all interpretability algorithms handle the same range of values, making it difficult to compare the explanations of different algorithms. Inspired by the majority voting principle in random forests, the voting approach is proposed. This function sorts the features by attribution in descending order and counts how often each feature is present among the $N$ most attributed features. Finally, the features appearing most frequently in the top $N$ are assumed to be the most important. The election of $N$ is a crucial but difficult task. If it is too small, it may be difficult to find a pattern amongst the top-ranked features, but if it is too large, features with very low attribution will be taken into account. Its main limitation is that it omits the attributions assigned to the features and does not consider either the contribution sign or the model accuracy.

\paragraph{Relative position} A similar approach, based on sorting the features by descendant attribution, is simply taking the relative position of each feature in the list. Feature positions then replace the attributions. Hence, this method ensures that the attributions are normalized in a common range: [1, number of features]. The main difference between this method and the previous ones is that, here, the lower the combined attribution, the more important the feature. Like voting, this function discards the actual magnitude of the feature attributions.

\paragraph{Other functions} In recent years, many studies about the application of ML and Deep learning (DL) models to the field of data fusion have been published \cite{Zamani2023,Rocken2024,Steyaert2023,Singh2023}. However, due to the complexity of the proposed models and the expensive computational requirements they could bring, it was decided not to include them in this study.

\subsection{Development of a novel consensus function}

The functions described in the previous section omit some crucial parameters to perform consensus accurately. To address these limitations, we developed a novel consensus function that incorporates these critical factors.

\paragraph{Identified challenges} To overcome the aforementioned issues, the proposed approach should address the main limitations identified.
\begin{enumerate}
\item Based on attributions. Some consensus functions, such as voting or the relative position, ignore the importance with which the input features contribute to the decision. Although they can be less computationally expensive, they overlook the magnitude of the contributions provided by interpretability algorithms.

\item Scale attributions. The interpretability approaches used in this work do not handle attributions in the same range, making comparing them difficult and unfair. In consequence, the proposed function will normalize all the attributions in the range [0, 1] using the min-max approach (Eq. \ref{eq:normalization}).

\begin{equation}
\label{eq:normalization}
attr_{scaled} = \frac{attr-min(attr)}{max(attr)-min(attr)}
\end{equation}

\item Distinguish global and local explanations. Global interpretability methods return a single attribution value per feature, which is inferred from the attributions of all the individual samples. On the contrary, local methods return a single value per sample. Therefore, if both values were combined into one single equation, local methods would have more importance simply because they contribute with more values. To overcome this problem, the attributions of the local methods have to be divided by the number of samples so that each local attribution contributes in the same proportion as the global ones.

\item Weight the errors. Local interpretability methods explain each sample individually. However, not all the samples are predicted with the same certainty. For example, a classification sample that is predicted with a probability of 0.5 is the result of a random decision. However, another sample, which is classified with a probability of 0.99, is a very reliable prediction whose explanation is of great interest. Similarly, the difference between the predicted value and the actual value in a regression problem is a measure of how reliable the prediction is. Hence, this can be an essential parameter when assessing local methods. The impact of class probability or regression error in model explanations should be represented by a correction factor.

\end{enumerate}

\paragraph{WISCA Formulation} WISCA (WeIghted Scaled Consensus Attributions) is the new consensus function proposed to avoid the above problems. It takes into account factors such as class probability (in classification problems), the difference between the actual and predicted value (in regression problems), and the different scales used by interpretability algorithms. In addition, it weights global and local interpretation methods fairly and equally. 

To define it formally, three cases are differentiated according to the type of problem and the type of interpretability algorithm used: i) global explanations (for classification and regression) (Eq. \ref{eq:wisca_global}); ii) local explanations in classification problems (Eq. \ref{eq:wisca_local_clf}); and iii) local explanations in regression problems (Eq. \ref{eq:wisca_local_reg}).

\begin{equation}
\phi(f)=\phi^{'}(f)\label{eq:wisca_global}
\end{equation}

\begin{equation}
\phi(f)=\sum_{s\in S}\frac{\phi^{'}(f)}{N}*\pi(s,m)\label{eq:wisca_local_clf}
\end{equation}

\begin{equation}
\phi(f)=\sum_{s\in S}\frac{\phi^{'}(f)*e^{-\alpha\left|y-\hat{y}\right|}}{N}\label{eq:wisca_local_reg}
\end{equation}

In equations \ref{eq:wisca_global}, \ref{eq:wisca_local_clf} and \ref{eq:wisca_local_reg}, $\phi^{'}(f)$ is the feature attribution scaled between 0 and 1 with the min-max algorithm, $m$ is the model, $f$ is the input feature, $S$ is the set of samples in the dataset, $s$ is the current sample, $N$ is the total number of samples in the dataset, $\pi(s,m)$ is the correction factor applied to classification problems and $e^{-\alpha\left|y - \hat{y}\right|}$ is the corrector factor used in regression problems.

\paragraph{Correction factor in regression} The function chosen to model the correction factor in regression problems (Eq. \ref{eq:wisca_local_reg}) measures the distance, in absolute value, between the value predicted by the model and the real value of the sample. Thus, when the distance between both values is very small, the correction factor increases, which implies keeping the attribution of the variable close to its original value. Conversely, when the distance between the two values is very large, the correction factor decreases and reduces the impact of the attribution on the consensus.

\begin{figure}[t]
\includegraphics[scale=.4]{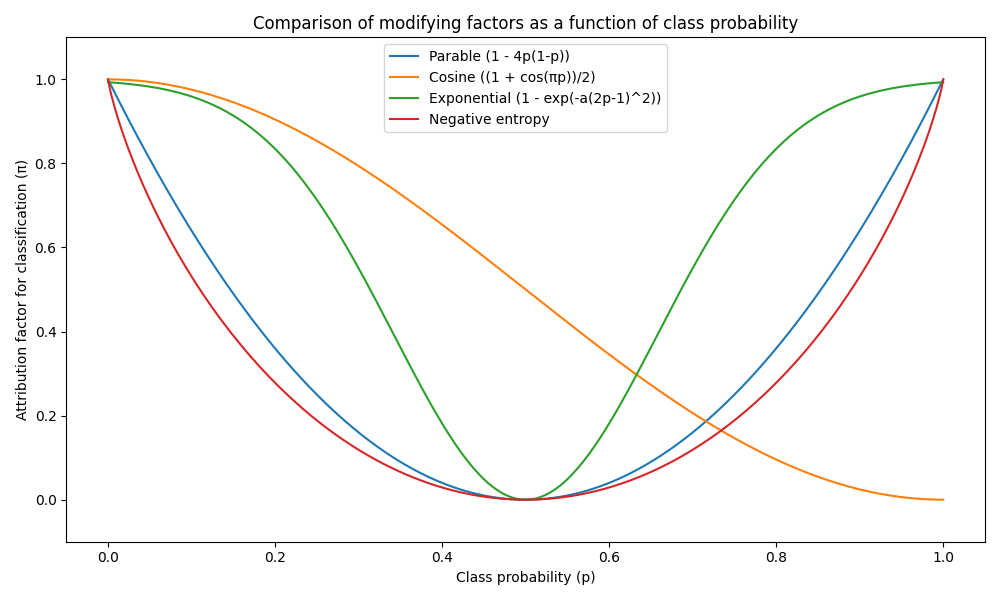}
\caption{Families of functions that can implement the classification correction factor.}\label{fig:alpha-factors}
\end{figure}

\paragraph{Alternatives for the classification correction factor} When implementing the correction factor in classification problems, a function is sought that reaches its maximum point when the input is 0 or 1 and the minimum at 0.5. These values correspond to a curve that is maximized when the sample has been predicted with absolute certainty or if the prediction has been totally wrong. However, when the uncertainty is maximized, the correction factor will cause the attribution to be minimized until it disappears completely. There are several families of functions that exhibit the desired behavior, as shown in Figure \ref{fig:alpha-factors}.

\begin{enumerate}
\item Parable. This function represents a smooth and symmetrical curve. It shows a rapid drop towards the minimum at p=0.5. It reaches its maximum at p=0 and p=1.

\begin{equation}
\pi(p) = 1 - 4p(1 - p)\label{eq:parable}
\end{equation}

\item Power. This function behaves similarly to the parabola, but does not touch exactly 1 at the ends. When n=2 it shows a sharper shape than the parabola. As its behaviour is nearly identic to the parable, consequently, it can be omitted.

\begin{equation}
\pi(p) = \left|2p - 1\right|^{2}\label{eq:power}
\end{equation}

\item Cosine. The cosine function shows a smooth and wavy behavior. It reaches perfect maxima at 0 and 1 and a minimum at 0.5. However, the transition is much slower than that of the parabola.

\begin{equation}
\pi(p)=\frac{1+cos(\pi p)}{2}\label{eq:cosine}
\end{equation}

\item Exponential. This function is displayed as an inverted bell. The fall speed is adjustable by means of a parameter ($a$). The higher $a$, the steeper the fall. At probabilities close to 0 and 1 it behaves more smoothly. However, we want our function to be more constant in its curvature.

\begin{equation}
\pi(p)=1-exp(-a(2p-1)^{2})\label{eq:exponential}
\end{equation}

\item Negative entropy. It behaves very similar to the parable function, but the curve is a little steeper. That makes the factor take longer to grow, i.e., intermediate between (0, 0.5) and (0.5, 1) the attributions become more similar because the correction factor varies less.

\begin{equation}
\pi(p)=1+p\ log_{2}p + (1 - p) log_{2}(1 - p)\label{eq:neg_entropy}
\end{equation}

\end{enumerate}

Among all the families of functions evaluated, the one that performs best as expected by WISCA are the parabolic functions. Consequently, this approach has been used to model the corrector function for classification problems.

\section{Results}
\label{sec:results}
\subsection{Model training}

\begin{figure}[t]
\includegraphics[scale=1]{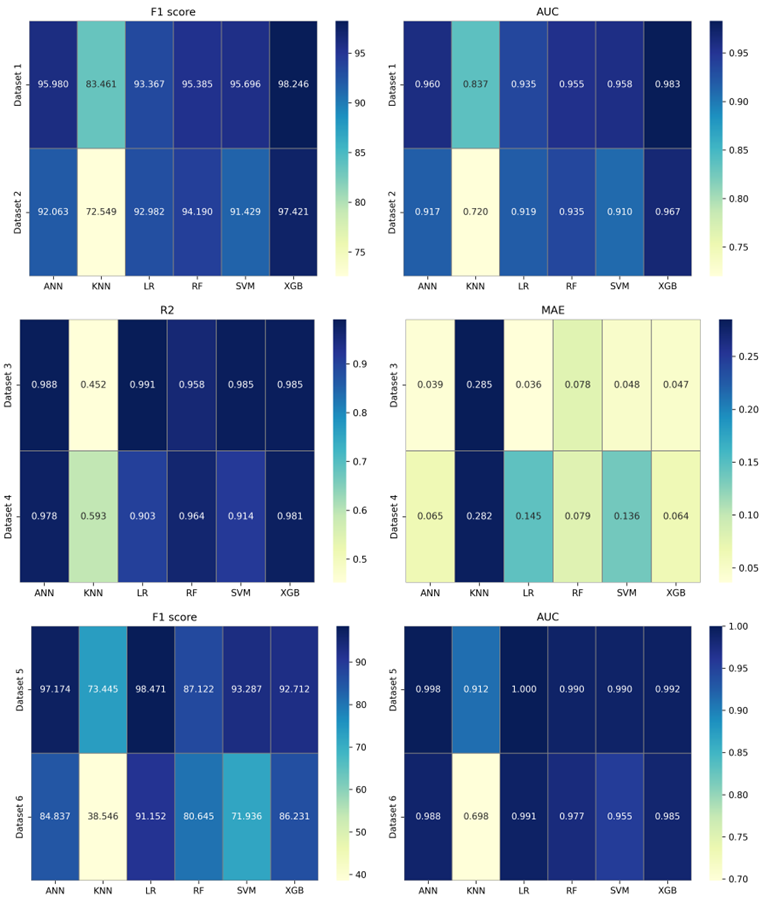}
\caption{Heatmaps summarizing model performance metrics across datasets.}\label{fig:model-training}
\end{figure}

To ensure robust results, each of the six models was trained ten times with each dataset by using the SIBILA tool \cite{Banegas2022}. The instances yielding the highest AUC (for classification) and R$^2$ (for regression) were selected for evaluation. Additional metrics, including F1-score and mean absolute error (MAE), were also collected. Figure \ref{fig:model-training} displays the main metrics of the six models after training.

\subsection{Consensus}

This section describes the experiments carried out to evaluate the accuracy of the consensus functions. The experiments aimed to assess how accurately each consensus function identifies the most relevant features in each dataset. Ideally, the features explaining the output should be the most attributed after consensus.

The consensus functions were tested on the six synthetic datasets, two for binary classification, two for multiclass classification and two for regression problems. The output of each dataset was generated manually using a different rule to test consensus functions in a variety of cases. Therefore, the features involved in the calculation of the output were known in advance and are summarized in Table \ref{tab:datasets}. 

To enable fair comparison, we defined a custom metric - termed the \textit{hit rate metric} - that calculates the precision of each function by assigning a weight to each feature that is considered relevant according to its position in the list of features ordered from highest to lowest attribution. Features that do not participate in the target calculation formula do not receive a score. According to this logic, if the $N$ variables used to generate the target are ranked highest after consensus, the function will obtain the highest score. As the position of these variables decreases and features introduced as noise are interspersed, the score goes down. The function assigns scores between 0 and 1, with 1 being the highest score and 0 the lowest. The mathematical formulation of the metric is:

\begin{equation}
P=\frac{\sum_{i=1}^{N}\frac{h(x_{i})}{i}}{\sum_{i=1}^{min(n,N)}\frac{1}{i}}\label{eq:hits_metric}
\end{equation}

\begin{figure}[t]
\includegraphics[scale=.5]{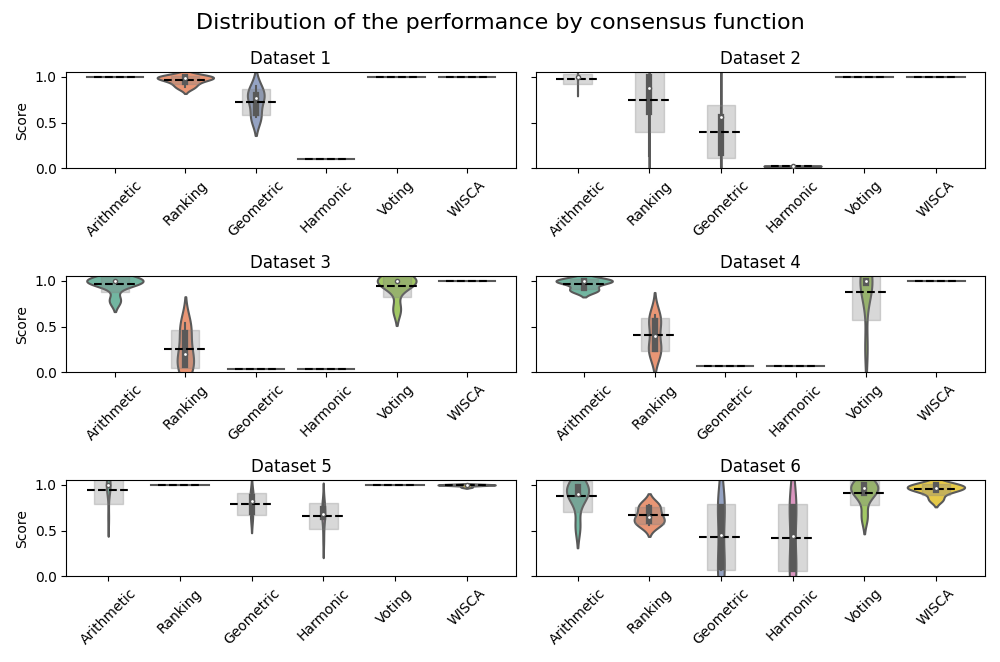}
\caption{Score of the consensus functions measured in terms of the hit rate metric.}\label{fig:hits-score-fn}
\end{figure}

\begin{figure}[t]
\includegraphics[scale=.75]{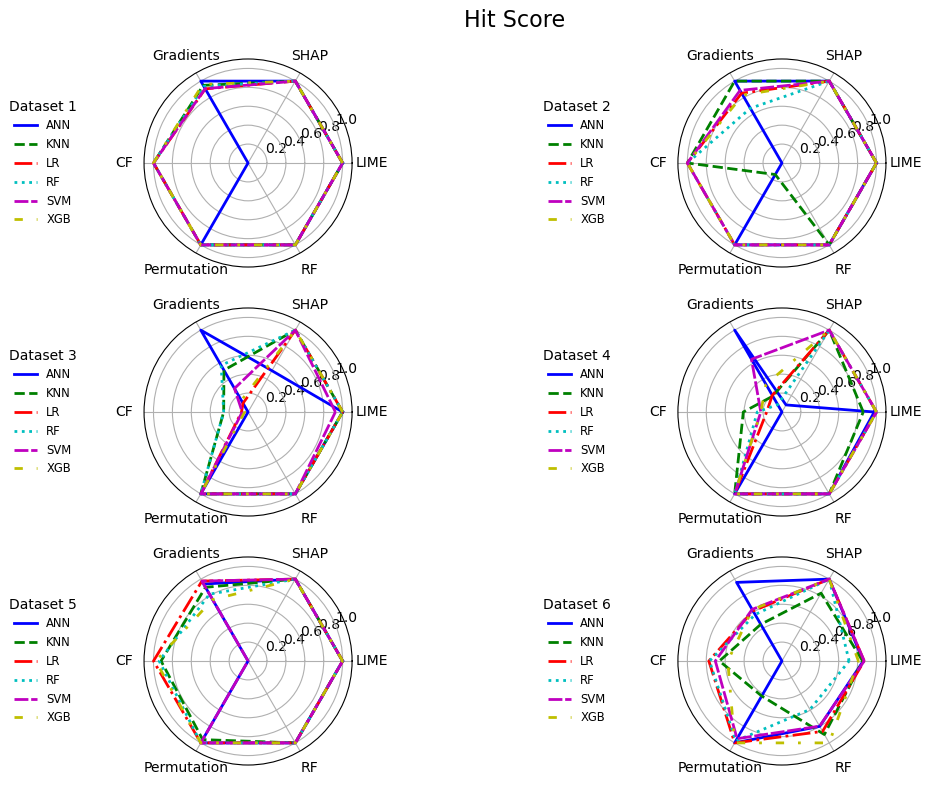}
\caption{Score of the interpretability algorithms measured in terms of the hit rate metric.}\label{fig:hits-score-alg}
\end{figure}

\begin{figure}[t]
\includegraphics[scale=.75]{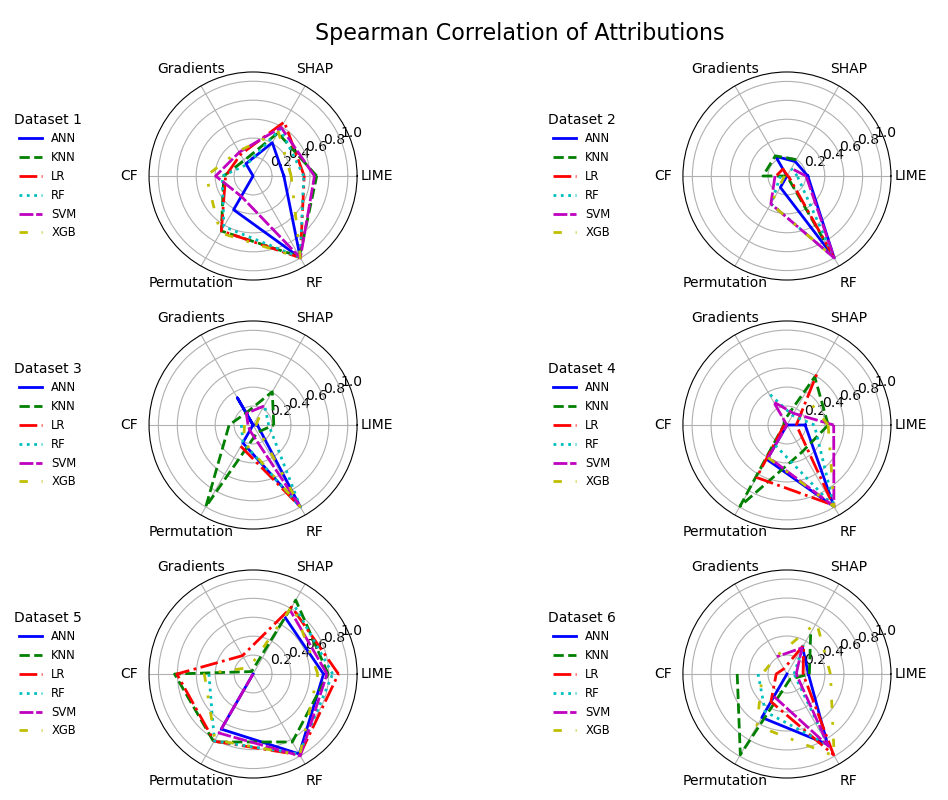}
\caption{Spearman correlation between WISCA and the interpretability algorithms.}\label{fig:spearman-wisca}
\end{figure}

\begin{figure}[t]
\includegraphics[scale=.75]{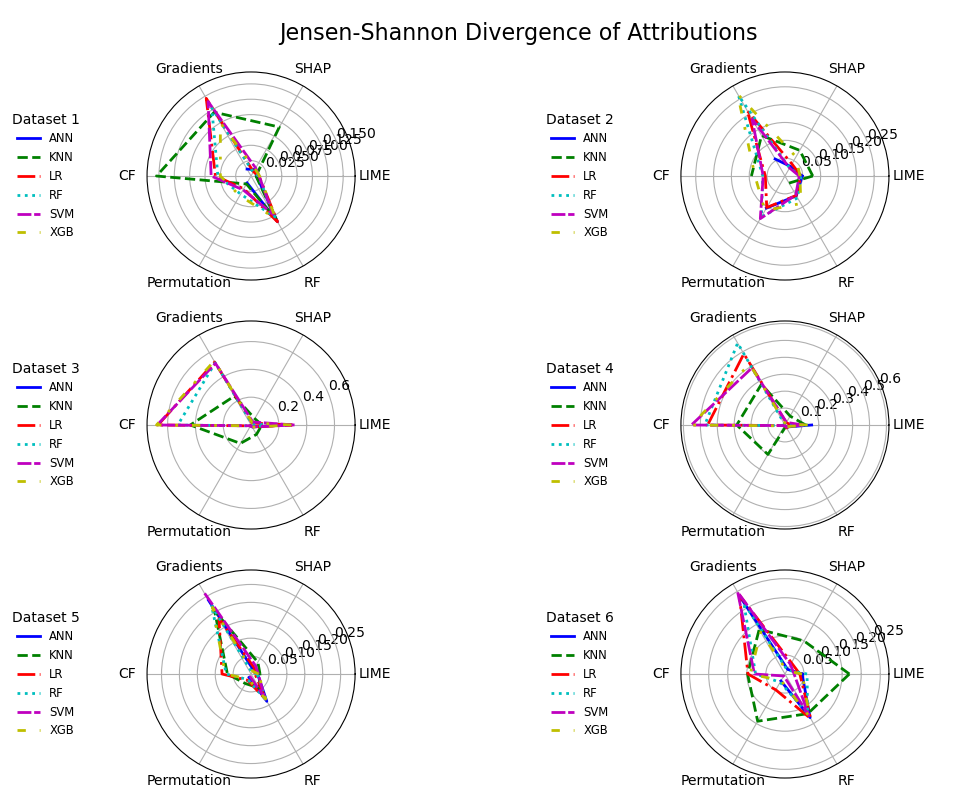}
\caption{Jensen-Shannon divergence between WISCA and the interpretability algorithms.}\label{fig:js-wisca}
\end{figure}

Figure \ref{fig:hits-score-fn} illustrates the comparative performance of consensus functions according to the custom hit rate metric. Higher values indicate better alignment with ground-truth relevant features. To evaluate WISCA’s ability to recover the most relevant features, we also computed the hit rate for the individual interpretability algorithms (Fig. \ref{fig:hits-score-alg}), along with Spearman correlation (Fig. \ref{fig:spearman-wisca}) and Jensen-Shannon divergence (Fig. \ref{fig:js-wisca}) between WISCA and each algorithm.

\begin{figure}[t]
\includegraphics[scale=.35]{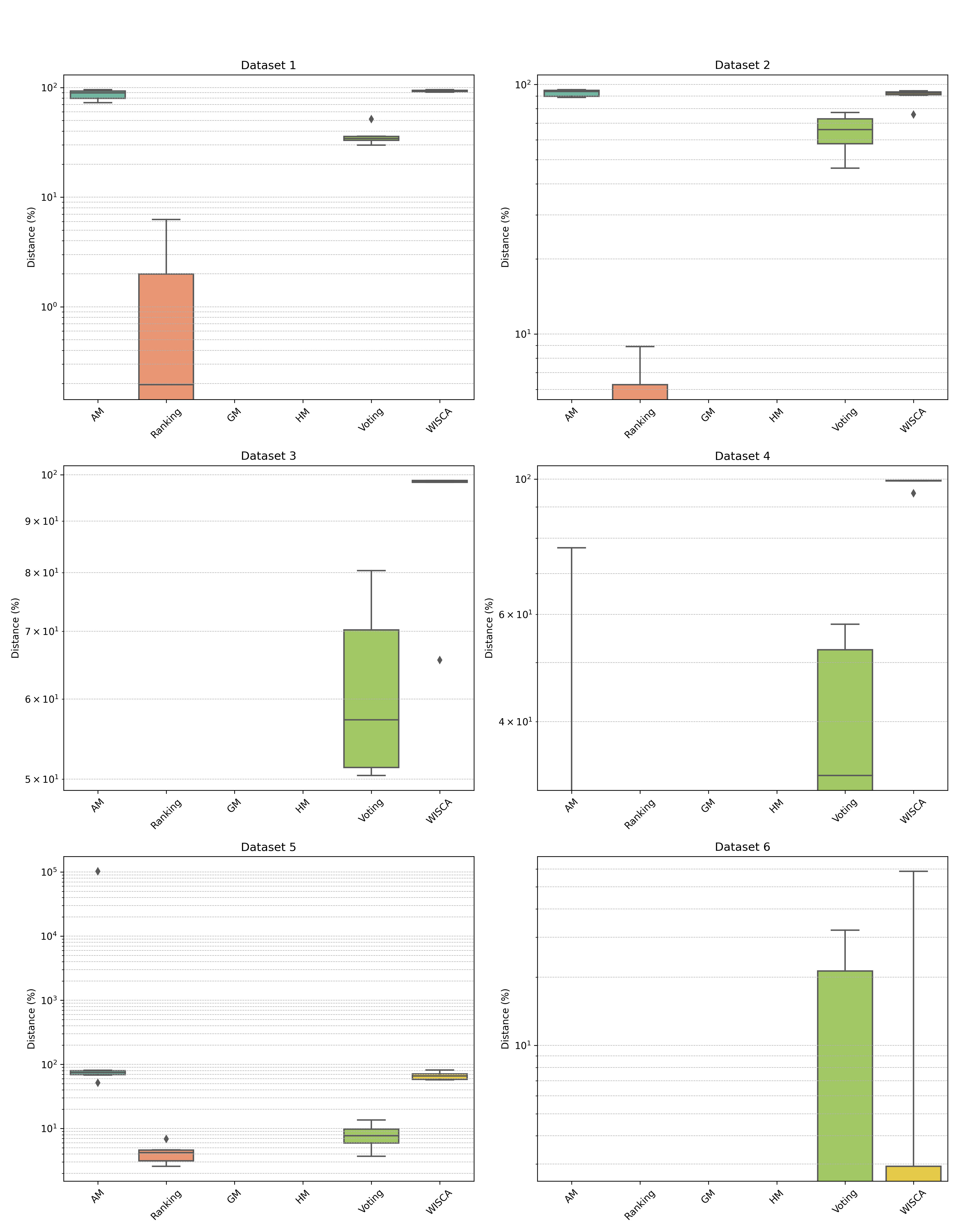}
\caption{Average distance between expected and non-expected features accross the datasets.}\label{fig:distance-fn}
\end{figure}

Finally, the ability of each consensus function to clearly distinguish between expected and non-expected variables has been analyzed. For this purpose, the distance between the two sets of variables, expected and nonexpected, was measured as:

\begin{equation}
    dist = \frac{\phi(EF_{min})-\phi(NEF_{max})}{max(\phi(EF_{min}), epsilon)} * 100
\end{equation}

where $\phi(EF_{min})$ is the minimum attribution assigned to an expected feature, $\phi(NEF_{max})$ is the maximum attribution assigned to a non-expected feature, and $epsilon$ is a constant, set to $10e^{-6}$, to avoid divisions by zero. The calculation was performed on a percentage basis to avoid biases due to the voting function. Since this function provides many more values, due to the fact that all local explanations vote equally, it would always be clearly the function that obtains the most difference. Figure \ref{fig:distance-fn} shows the average distance of each function accross the six datasets.

\section{Discussion}
\label{sec:discussion}
\subsection{Assessment of consensus functions}

Analyzing the functions for each group of datasets (Fig. \ref{fig:hits-score-fn}) shows that in the binary classification datasets (datasets 1 and 2), voting and WISCA always find all the expected variables in the first positions. The arithmetic mean also performs very well, although slightly worse than the others in the black box model (ANN).
In the multiclass classification cases there is more variety of results. Even so, voting still gives the expected result in dataset 5, closely followed by WISCA. On the other hand, in dataset 6 it is WISCA that slightly outperforms the voting function. Although the mean attribution is very similar in both functions, the standard deviation of WISCA is clearly lower, suggesting more compact attributions.
Finally, in the regression datasets (datasets 3 and 4) WISCA is the only function that manages to find all the expected features with all the models. While the arithmetic mean has a similar behavior in both datasets, the voting clearly worsens in dataset 4. In fact, the poor performance of the voting function is repeated in datasets 3, 4 and 6, exactly in datasets where some model, namely KNN, performed poorly during training. Therefore, it appears that this function is very sensitive to model accuracy.
In summary, although the arithmetic mean and voting perform well, they are outperformed by WISCA in the vast majority of cases. The latter finds the expected results in 4 of the 6 datasets, and is still the best (without being perfect) in another. On the other hand, the geometric mean and harmonic are very poor choices for computing consensus, while the ranking function barely performs accurately in a few cases.

It is also interesting to investigate why some functions perform so poorly in the consensus. The harmonic mean is defined as the reciprocal of the arithmetic, which already gives an intuition that if one gives good results, the other will give poor results. In the case of the harmonic mean, the greater the weight assigned to the variable, the less it contributes to the decision. This is exactly the opposite of what is expected and the reason why the harmonic mean is not a good alternative for implementing consensus.
For its part, the geometric mean systematically fails as a consensus function of attributions for several reasons. First, the geometric mean cancels out any explanation that contains a null attribution (or very close value), because if only one of the attributions is 0, then the product of all of them will also be 0. In the context of attributions, it is common that many variables do not contribute information to the explanation, for example, the variables that make up the noise usually have very low attributions. When using the geometric mean, these variables drag the entire consensus down to 0, masking the signal of the really relevant variables. Second, many interpretation techniques (e.g., integrated gradients, counterfactuals) return positive and negative attributions, signaling favorable or unfavorable contribution to the prediction. The geometric mean cannot handle negative values. Furthermore, the geometric mean does not preserve the same scale as the original values nor does it maintain additive orderings. In contrast, sum-based functions (arithmetic mean, WISCA, etc.) do maintain proportionality, which is key for comparing relative importance.
Finally, consensus ranking only considers the relative position of the features in the ordered list of variables by descending attribution. Unfortunately, some algorithms (e.g. feature permutation importance) assign a weight of 0 to many variables, which makes the position of the variables in the ranking dependent on the sorting algorithm or a secondary tie-breaker criterion. This can clearly affect the weight with which the features contribute to the consensus and consequently lead to unreliable results.

\subsection{Evaluation of WISCA}

Figure \ref{fig:hits-score-alg} shows the scores obtained individually by the interpretability algorithms in their original explanations, i.e., without using any consensus, when applying the hit rate metric. The best results are obtained when the values are close to 1 (the outer line) because they represent the algorithms that best interpret each model. This information will be useful to contrast with the following figures.

Spearman's correlation tells us whether the order of attributions is preserved, i.e., whether there is a relationship between the original and consensus attributions. To measure it, we have taken the original averaged attributions. In the global interpretability methods, these are the attributions returned directly; whereas, in the local methods, it is calculated as the average of the attribution of each variable across all samples. This information has been correlated with the attributions calculated by WISCA. It is expected that, in each dataset, the algorithms with the highest Spearman correlation correspond to those with the highest hit rate (Fig. \ref{fig:hits-score-alg}). It is observed that in datasets 1, 2 and 5 the highest correlation is with the RF approach; and in datasets 3, 4 and 6 with Permutation and RF. Figure \ref{fig:spearman-wisca} shows that the algorithms indicated here are almost always the ones with the highest hit rate. This means that WISCA has the ability to assign attributions that closely resemble those of the algorithm that best explains each model.

Next, the Jensen-Shannon divergence has been calculated with the same data as the Spearman correlation, that is, between the consensus and original attributions. Paying attention to Figure \ref{fig:js-wisca}, it can be seen that the Jensen-Shannon divergence between WISCA and the original attributions is usually very close to zero. This means that the distribution of the data is very similar between the consensus and original values. That confirms that WISCA is able to obtain very similarly distributed attributions to the algorithm that best explains each model.

\subsection{Linear models vs WISCA}

\begin{figure}[t]
\includegraphics[scale=.5]{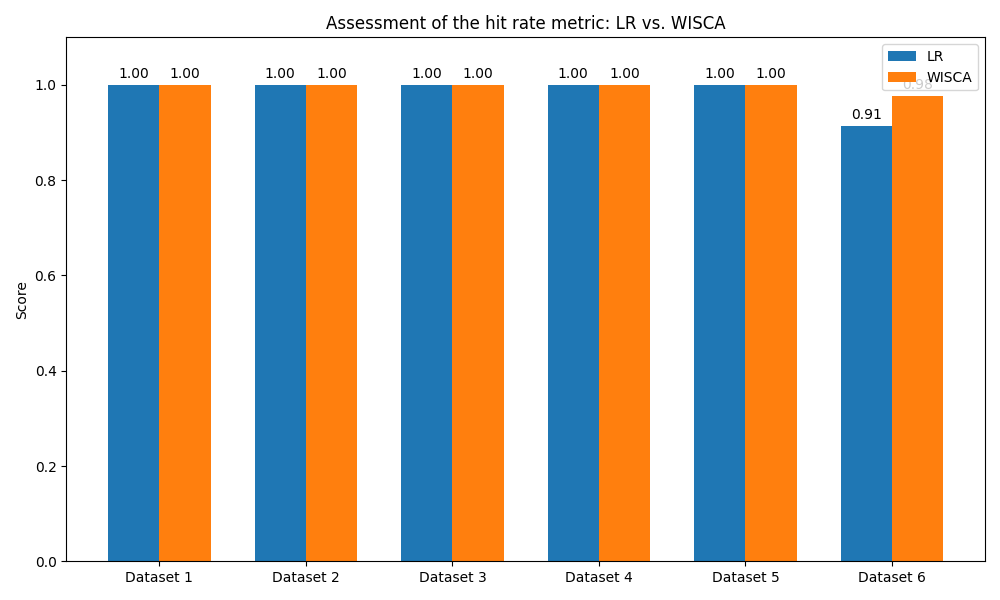}
\caption{Hit rate score of LR and WISCA.}\label{fig:wisca-vs-lr}
\end{figure}

Of particular interest is comparing WISCA against a linear model (LR), also. As the linear models trained reached high accuracy, it was expected they yelded accurate explanations. To perform the comparison, the new hit rate metric was calculated for both WISCA and LR. In order to calculate the hit rate metric for the linear model, the features were sorted by descending order of the computed coefficients. Based on that, the importance of the features calculated by LR and WISCA is compared. Figure \ref{fig:wisca-vs-lr} confirms that WISCA behaves similarly to a linear model. Even in Dataset 6 our function outperforms the linear model, which obtained an F1 score = 91.152 and AUC = 0.991.

This finding proves the ability of WISCA to identify the most relevant features. In addition, the explanations created with WISCA are supported by several interpretability algorithms, making them more robust and consistent. 

\subsection{Real datasets}

\begin{figure}[t]
\includegraphics[scale=.6]{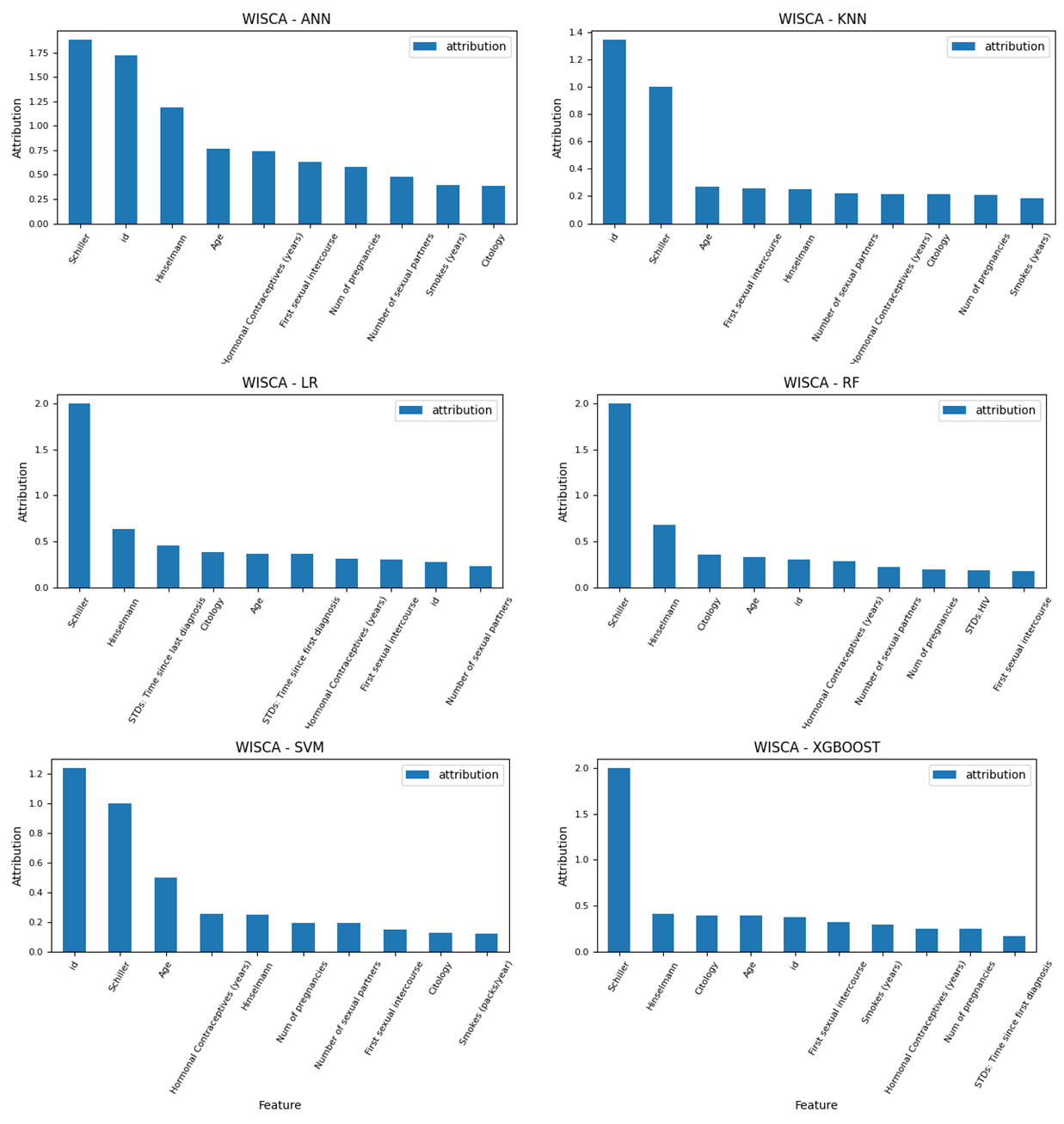}
\caption{Consensus explanations returned by WISCA on the cervical cancer risk dataset.}\label{fig:toy-clf-bin}
\end{figure}

\begin{figure}[t]
\includegraphics[scale=.6]{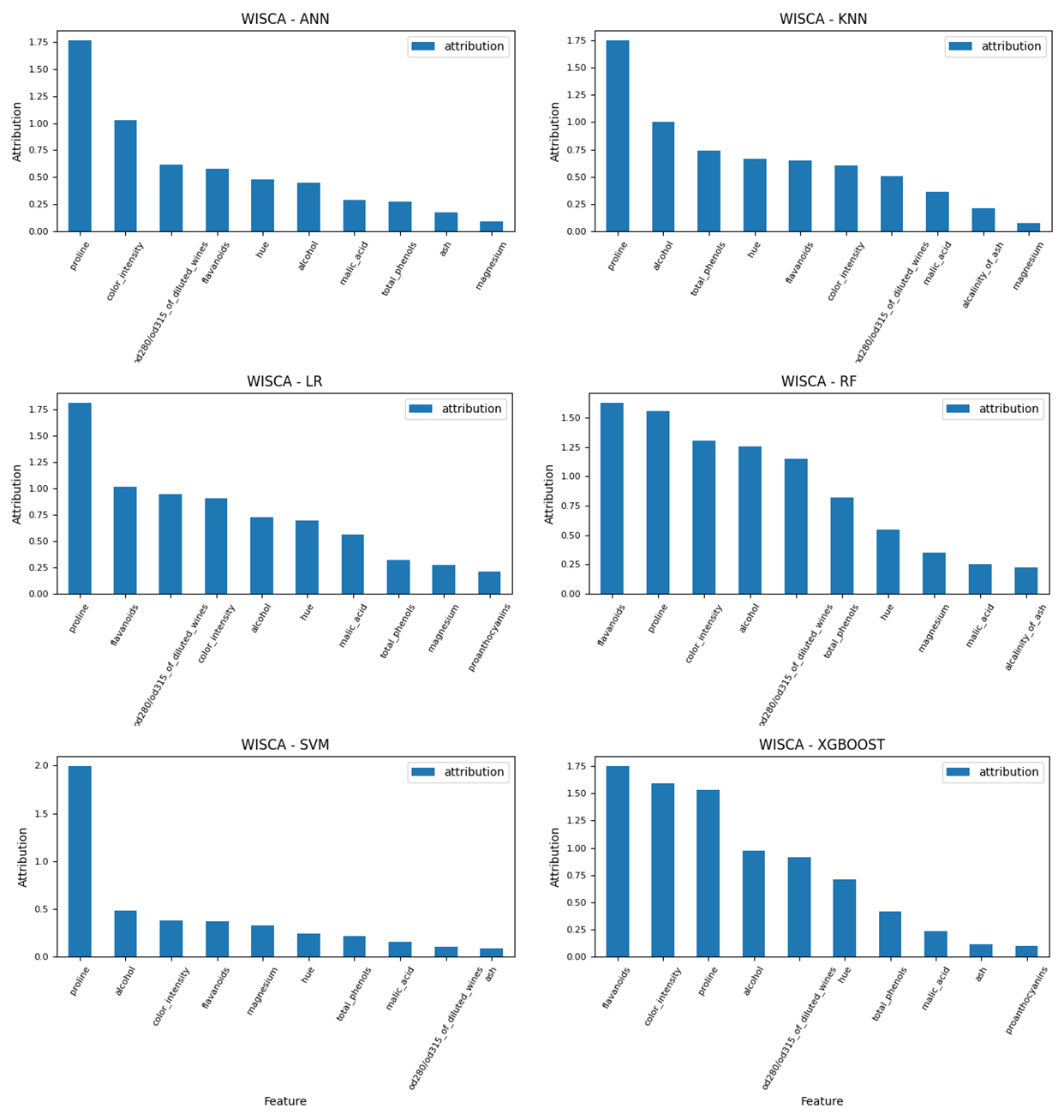}
\caption{Consensus explanations returned by WISCA on the wine dataset.}\label{fig:toy-clf-multi}
\end{figure}

\begin{figure}[t]
\includegraphics[scale=.6]{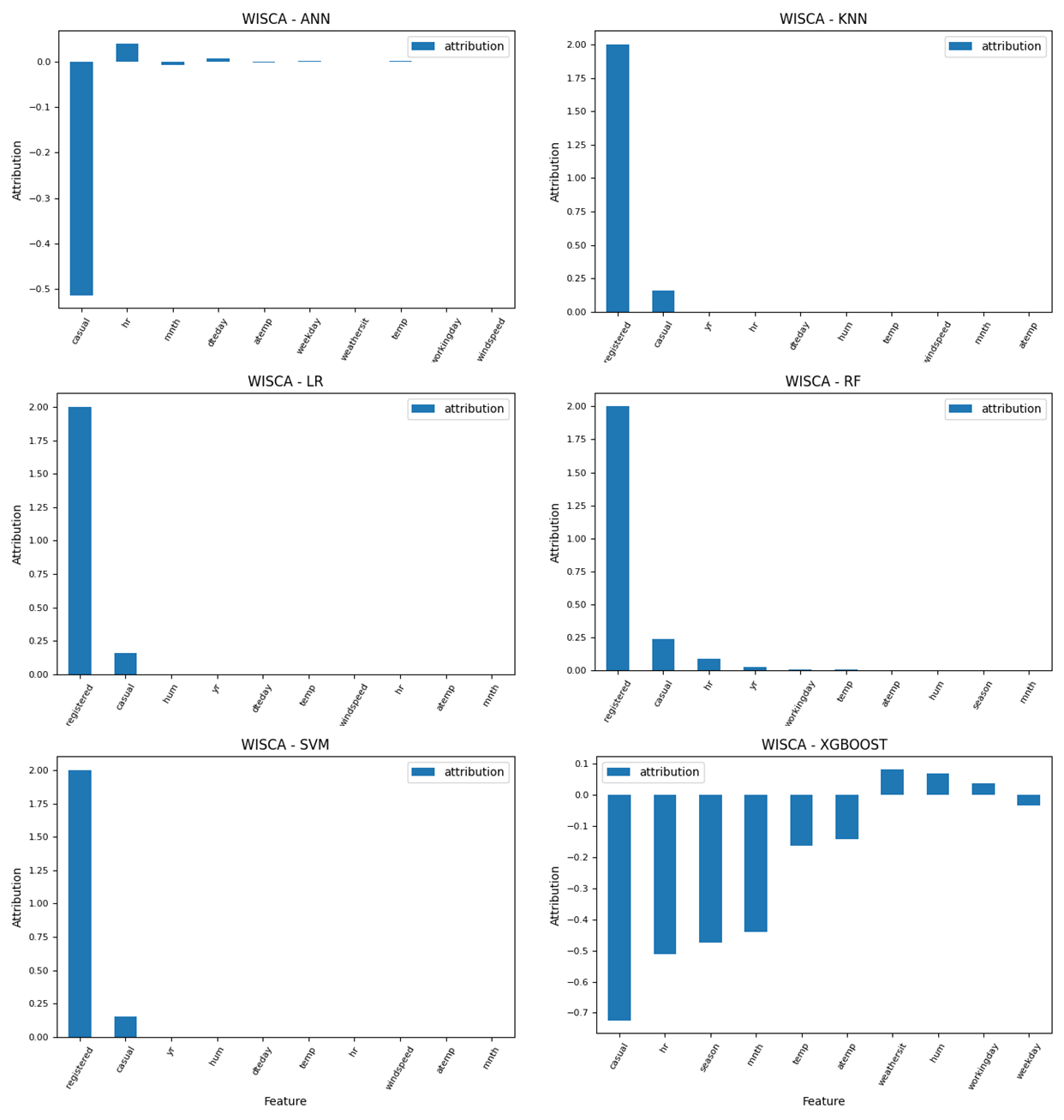}
\caption{Consensus explanations returned by WISCA on the bike rental dataset.}\label{fig:toy-reg}
\end{figure}

WISCA has also been tested on toy datasets. Three public datasets have been selected: cervical cancer risk \cite{cervical_cancer_(risk_factors)_383} for binary classification, origin of wine classification \cite{wine_109} for multiclass classification, and bike rental \cite{bike_sharing_275} for regression. Figures \ref{fig:toy-clf-bin}, \ref{fig:toy-clf-multi} and \ref{fig:toy-reg} show the explanations obtained through WISCA in the three cases.

In the cervical cancer risk three models (LR, RF and XGB) clearly highlighted the importance of the \textit{Schiller} feature. KNN and SVM also highlighted the importance of \textit{id}. Finally, the ANN model is explained by those two features, but their importance is closer to the rest of the features. The Schiller test is a gynecological study that applies iodine to the cervix to detect cellular alterations. It is performed during colposcopy. The Schiller test helps to define the boundaries between the epithelium and the lesion. Therefore, it is logical that the result of this test is considered an expected marker for detecting this type of cancer. On the other hand, \textit{id} is the sample identifier, which, as expected, cannot be considered the only explanation of a model.

Regarding the wine dataset, in four of the six models \textit{proline} was the most ranked feature. Proline is an amino acid found naturally in grapes that has recently been shown to enhance the viscosity, sweetness, and flavor of red wine. Grapes high in proline are associated with warmer climates and riper grapes. As far as we are concerned, the identification of proline as a marker of wine origin is totally accurate. The other two models (RF and XGB) ranked \textit{proline} among the three most relevant features too. In both cases, a small gap can be observed between \textit{proline} and the next ranked feature. 

Finally, in the bike rental dataset, four out of six models can be undoubtedly explained with the \textit{registered} feature. This means that the number of bicycle rentals increases as the number of registered users increases. On the other hand, the ANN and XGB models use the opposite reasoning to explain their behaviour. They assign a very negative attribution to the \textit{casual} feature, which means that the least number of unregistered users, the more rentals happen. In short, WISCA was able to logically explain the three tested datasets.

\section{Conclusions and future works}

Understanding how black-box AI models work is essential. Various approaches have emerged to shed light on their decision-making processes in recent years. While these methods aim to quantify the contribution of each input feature, they employ different strategies. Addressing the disparities among these algorithms and reaching a consensus could offer a promising solution. Consensus can take different forms, such as averaging, considering the relative importance of features, or tallying their frequency among the most relevant ones.

In this study, we evaluated five consensus functions using six synthetic datasets in different types of problems. Our findings revealed crucial limitations across all functions, such as the lack of information about the models' accuracy, the problems dealing with missing or negative attributions, the difficulties in attributing significantly relevant features, and the lack of a common scale for all the attributions. Consequently, we proposed a novel consensus function, WISCA, that considers factors like class probability and attribution normalization to comprehensively explain model outcomes. WISCA surpassed others by effectively identifying features used in generating outputs in most cases. These results underscore the importance of considering multiple factors for accurate consensus. An effective consensus function is crucial for explaining model predictions, particularly in critical fields like medicine. Nevertheless, in real-world datasets where the importance of the features is unknown, human validation is necessary to ensure the explanations are valid.

Future research will concentrate on applying consensus to more real-world datasets with unknown internal structures. In addition, WISCA could be extended to take into account the predicted class in classification problems. That way, samples accurately predicted would contribute to the explanation more than the wrongly predicted ones. Moreover, expanding the range of consensus functions is a promising avenue. Leveraging ML models to predict feature attributions based on existing interpretations could offer an alternative approach. This entails developing a novel model using a dataset derived from previous interpretations to predict feature attributions effectively.





\begin{acknowledgement}[title={Acknowledgments}]
This work has been funded by grants from the European Project Horizon 2020 SC1-BHC-02-2019 [REVERT, ID:848098]; Fundación Séneca del Centro de Coordinación de la Investigación de la Región de Murcia [Project 20988/PI/18]]; and the Spanish Ministry of Economy and Competitiveness [CTQ2017-87974-R]. Supercomputing resources in this work were supported by the Poznan Supercomputing Center’s infrastructures, the e-infrastructure program of the Research Council of Norway, and the supercomputing centre of UiT—the Arctic University of Norway, the Plataforma Andaluza de Bioinformática at the University of Málaga, the supercomputing infrastructure of the NLHPC (ECM-02, Powered@NLHPC), and the Extremadura Research Centre for Advanced Technologies (CETA-CIEMAT), funded by the European Regional Development Fund (ERDF). CETA-CIEMAT is part of CIEMAT and the Government of Spain.
\end{acknowledgement}

\begin{funding}
This work has been funded by grants from the European Project Horizon 2020 SC1-BHC-02-2019 [REVERT, ID:848098]; Fundación Séneca del Centro de Coordinación de la Investigación de la Región de Murcia [Project 20988/PI/18]; and the Spanish Ministry of Economy and Competitiveness [CTQ2017-87974-R].
\end{funding}

\bibliographystyle{infor}
\bibliography{biblio}
\vfill

\begin{biography}\label{bio1}
\author{A.J. Banegas-Luna} is an Associate Professor in Computer Science at Universidad Católica de Murcia (UCAM), Spain. He earned his Ph.D. in Computer Science from UCAM in 2019, specializing in the application of high-performance computing (HPC) to biological and chemical contexts. His research focuses on computer-aided drug discovery and the use of artificial intelligence techniques for personalized therapy development. With over 15 years of experience in large-scale IT projects for public organizations and European institutions, he has expertise in web development across various programming languages, frameworks, and databases. Dr. Banegas-Luna is a member of the Structural Bioinformatics and High Performance Computing Research Group (BIO-HPC) at UCAM, where he contributes to multidisciplinary research in bioinformatics applications, supercomputing, and the discovery of bioactive compounds. He has published many papers in indexed journals and has participated in international congress and invited talks.
\end{biography}

\begin{biography}\label{bio2}
\author{H. Pérez-Sánchez} is a computational chemist with an international career across Spain, France, and Germany, supported by 11 fellowships, including two Marie Curie. I currently lead the Structural Bioinformatics and High Performance Computing group at UCAM (bio-hpc.eu), where we integrate structural bioinformatics, HPC, and physical chemistry to develop tools for molecular modeling and data interpretation. Many of these tools run on supercomputers or GPUs, are freely accessible online, and some have been commercialized with industry partners.
His work applies to protein and DNA modeling, analysis of structural data (mutations, complex formation, ion channels, food science), and drug discovery (cancer, antivirals, anticoagulants). He has supervised 11 PhD theses (3 industrial) and lead over 30 research projects, securing €860k in direct and €6.1M in indirect funding. His output includes 198 JCR papers (over 100 in Q1), 48 peer-reviewed conference papers, 11 patents (5 licensed), and 6 industry contracts. He holds ANECA accreditations for Profesor Contratado Doctor and Profesor Titular, and the I3 certification. He also hosts a Spanish-language research podcast with over 330 episodes: \url{anchor.fm/horacio-ps}.
\end{biography}

\begin{biography}\label{bio3}
\author{C. Martínez-Cortés} is a researcher at the Fundación Universitaria San Antonio (UCAM), where he is part of the Structural Bioinformatics and High Performance Computing research group (BIO-HPC) at UCAM HiTech, Sport \& Health Innovation Hub. He holds a PhD in Computer Science from the University of Murcia, as well as Master’s and Bachelor’s degrees in Computer Engineering. His research spans machine learning, interpretable artificial intelligence, and virtual screening for drug discovery. He has contributed to several open-source tools, including MetaScreener and Sibila, and has co-authored various publications and patents in the area of bioinformatics and computational drug design. He is actively involved in the EU-funded REVERT project, which focuses on personalized medicine for colorectal cancer patients.
\end{biography}

\end{document}